\documentclass[snmathphys]{snjnl}

\jyear{2021}%

\theoremstyle{thmstyleone}%

\theoremstyle{thmstyletwo}%

\theoremstyle{thmstylethree}%

\begin{document}
	
	\title[Article Title]{A Random Projection $k$ Nearest Neighbours Ensemble for Classification via Extended Neighbourhood Rule}
	
	\author[1]{\fnm{Amjad} \sur{Ali}}\email{Amjad.ali@awkum.edu.pk}
	
	\author[1]{\fnm{Muhammad} \sur{Hamraz}}\email{mhamraz@awkum.edu.pk}
	
	\author[1]{\fnm{Dost} \sur{Muhammad Khan}}\email{Dostmuhammad@awkum.edu.pk}
	
	\author[2]{\fnm{Wajdan} \sur{Deebani}}\email{wdeebani@kau.edu.sa}
	
	\author*[1]{\fnm{Zardad} \sur{Khan}}\email{zkhan.essex@gmail.com}

	\affil*[1]{\orgdiv{Department of Statistics}, \orgname{Abdul Wali Khan University}, \orgaddress{\city{Mardan}, \postcode{23200}, \state{KPK}, \country{Pakistan}}}
	
	\affil[2]{\orgdiv{Department of Mathematics}, \orgname{College of Science and Arts, King Abdul Aziz University}, \orgaddress{\city{Rabigh}, \country{Saudi Arabia}}}
	

	\abstract{Ensembles based on $k$ nearest neighbours ($k$NN) combine a large number of base learners, each constructed on a sample taken from a given training data. Typical $k$NN based ensembles determine the $k$ closest observations in the training data bounded to a test sample point by a spherical region to predict its class. In this paper, a novel random projection extended neighbourhood rule (RPExNRule) ensemble is proposed where bootstrap samples from the given training data are randomly projected into lower dimensions for additional randomness in the base models and to preserve features information. It uses the extended neighbourhood rule (ExNRule) to fit $k$NN as base learners on randomly projected bootstrap samples. Several benchmark datasets are used for assessing the performance of the proposed method in comparison with other state-of-the-art procedures. Classification accuracy, Brier score and Cohen’s kappa are used as comparison metrics. The proposed RPExNRule ensemble and other classical procedures are also compared via simulated data. Based on benchmarking and simulation study, the results given in the paper reveal that the proposed method performs better than the others in the majority of the cases.
	}
	
	\keywords{Random Projection, Dimension Reduction, Classification, Extended Neighbourhood Rule, $k$NN Ensemble}
	
	\maketitle
	
	\section{Introduction}
	\label{introduction}
	
	Classification is one of the most important topics used in supervised learning problems. It classifies unseen observations to their respective classes based on a group of related features in the training data. $k$ nearest neighbours ($k$NN) \cite{cunningham2021k, altman1992introduction, hastie2009friedman} is one of the top-ranked classification procedures \cite{wu2008top}. This method was developed by Fix and Hodges in \cite{fix1951discriminatory} and then used for pattern recognition by Cover and Hart in \cite{cover1967nearest}. It is one of the most simple classification procedures which gives promising results in the case of big datasets \cite{hand2007principles, amal2011survey, kulkarni2013introspection, abbasifard2014survey}. It should be the first choice to use when there is a small amount or no prior information available about the data distribution. The main idea behind the $k$NN method is to classify an unseen observation into the most frequent class by using majority voting in the labels of the identified $k$ nearest sample points in the feature space. As the $k$NN procedure is one of the top ten fundamental and non-parametric data mining procedures \cite{wu2008top} that gives efficient results on several benchmark datasets. However, it may be affected by several data-related problems, like the existence of noise and non-informative features in the data. The classical $k$NN have been modified for further improvement, such as weighted $k$ nearest neighbours (W$k$NN) \cite{bailey1978note} and condensed nearest neighbours (CNN) \cite{alpaydin1997voting, zhai2011condensed, gowda1979condensed}.
	
	Ensembles of base $k$NN learners in conjunction with randomized sampling procedure(s) bring diversity to the classifier and make the models more stable and flexible. Randomization is injected into these procedures by taking random bootstrap samples from the data points and/or random subsets of features from the feature space involved in the data for constructing the base $k$NN models. This makes the model more diverse and reduces the chance of the same errors occurring repeatedly \cite{bay1999nearest, domeniconi2004nearest, garcia2009boosting}. Therefore, these methods are more stable and reliable as compared to the single learner classifiers. Moreover, random subsets of the features are also used for dimensionality reduction which minimizes the execution time. Examples of the $k$NN based ensembles are bootstrap aggregated $k$NN \cite{steele2009exact}, random $k$NN \cite{li2014random}, ensemble of subset of $k$NN \cite{gul2018ensemble}, weighted heterogeneous distance Metric \cite{ZHANG201913}, ensemble of random subspace $k$NN \cite{rashid2021random}, optimal $k$NN ensemble \cite{ali2020k}, etc. Almost all the $k$NN ensembles use majority voting in the class labels of the nearest instances of the test observation to estimate its class in each base model. The final prediction is obtained by using the second round of the majority voting in the results given by all the base learners. As the $k$NN based ensembles combine a large number of base learners, where each model is constructed on a bootstrap sample in conjunction with a subset of features. Therefore, many of these learners are fitted on irrelevant features and several influential features are ignored. Moreover, such ensembles do not perform well where the test observations follow a pattern of the nearest training observations that lie on a certain path beyond the sphere of the neighbourhood. Due to these problems, the conventional ensembles become weak and hence, give poor prediction accuracy. 
	
	A novel random projection extended neighbourhood rule (RPExNRule) for the $k$NN ensemble is proposed in this paper. The extended neighbourhood rule (ExNRule) \cite{ali2022k} is used for $k$NN as base learners. To construct each base model a bootstrap sample is drawn from the training data with the full feature space and then randomly projected into a reduced dimension. The predicted class of an unseen observation is obtained using majority voting in the results given by the base learners. For assessing the performance of the novel ensemble, 15 benchmark problems are used. The classification accuracy, Brier score and Cohen's kappa are used as metrics of performance. The $k$ nearest neighbour ($k$NN), weighted $k$ nearest neighbours (W$k$NN), random $k$ nearest neighbour (R$k$NN), random forest (RF), optimal trees ensemble (OTE) and support vector machine (SVM) are used as the competitors. For further evaluation, boxplots of the results are also created to show the precision of the performance.
	
	The remaining article is organized as follows. Section \ref{litrature} gives detailed literature related to the proposed work. A detailed description of the proposed method, its mathematical background and its algorithm are provided in Section \ref{method}. Section \ref{experiment} gives details of the datasets, experiments and discussion on the results. The paper is concluded in  Section \ref{conclusion}.
	
	\section{Related works}
	\label{litrature}
	
	The $k$ nearest neighbours ($k$NN) classifier is one of the top ten data mining procedures \cite{wu2008top} that gives efficient results in many real life problems. The classical $k$NN procedure gives equal weights to the observations determined in the neighbourhood of an unseen sample point. A weighted $k$ nearest neighbours (W$k$NN) algorithm is proposed by \cite{bailey1978note}. The weights are assigned based on the distance of the nearest observations from the unseen sample point. The W$k$NN procedure gives promising prediction results on several benchmark problems as compared to the classical $k$NN method. However, W$k$NN procedure uses all sample points in the training data which makes it a global method and time consuming. Authors have proposed several methods for the purpose to reduce the training time, such as condensed nearest neighbour (CNN) techniques \cite{alpaydin1997voting, gowda1979condensed}. The CNN procedures remove identical observations from the training data as these sample points have no extra information to improve the results of the model. However, these methods might ignore the sample points which lie on the boundary and depend on data order. \cite{gates1972reduced} proposed the reduced nearest neighbor (RNN) rule, which is a modified version of the CNN rule. This method removes instances and minimizes the training dataset. However, RNN procedure is computationally complex and laborious. The fast condensed nearest neighbor (FCNN) rule proposed by \cite{angiulli2005fast}. This method allows selecting observations that are much close to the decision boundary. The FCNN has a very small amount of quadratic complexity and is independent of the observations order. Another improvement have been made by \cite{guo2003knn}, using a model based $k$NN method. A function is fitted on the training data to estimate the test sample points. This method also reduces the training data size like CNN and RNN methods and provides promising prediction accuracy. However, this method avoids the marginal data points which lies farther along in the space. A clustered $k$NN (C$k$NN) is introduced by \cite{yong2009improved} which is one of the more robust methods and reducing the problem of unequal distribution of training sample points. The C$k$NN method determines the nearest instances in the clusters. To determine the threshold used in this technique is difficult task, to compute distances among the observations within different clusters. The selection of the optimal $k$ value is also unknown. Another method called modified $k$ nearest neighbour (M$k$NN) is introduced by \cite{parvin2008mknn} which uses weights and validity of the training data points to estimate the response value of a test observation. \cite{sproull1991refinements} introduced a $k$ dimensional tree nearest neighbour ($k$DTNN) technique, which splits the training data into sub-planes to orchestrate multidimensional samples. The $k$DTNN technique produces a tree that is balance, simple and easy to interpret. However, this method requires intensive search to split the instances blindly into sub-planes. Moreover, $k$DTNN involves computational complexity and might miss the data pattern. A sample reduction procedure for regression and classification proposed in \cite{an2022data}, inspired by man social behavior. Other improved ensembles can be found in\cite{tahir2007simultaneous, rohban2012supervised, chen2016effectively, xu2019knn, yuan2021novel, sun2021density, zhang2022drcw}.
	
	Several ensembles based on $k$NN methods are also proposed in literature, that make the classifiers more efficient and diverse. A bootstrap aggregation (bagging) is one of the basic ensemble techniques proposed by \cite{breiman1996bagging}. It constructs a new model to predict the exact bootstrap average of several base learners \cite{caprile2004exact, zhou2005adapt, zhou2005ensembling}. Bagging provides a basic platform for several state-of-the-art ensemble methods. Ensemble techniques based on bagging are constructed by using a large number of bootstrap samples drawn from training data. A base classifier is fitted on each sample and used to predict test data. The final prediction is obtained by using majority voting on the results given by the base learners \cite{breiman1996bagging}. The standard bagging procedure is further improved by \cite{steele2009exact}. Steel extended the exact bagging approach to sub-sampling aggregation with and without replacement techniques. Several ensembles methods have been introduced in literature which use both bagging and subsets of features to fit base $k$NN classifiers \cite{li2011random, li2014random, gul2018ensemble, gu2018random}. Several authors have also introduced $k$ value optimization based ensemble methods \cite{grabowski2002voting, zhang2017efficient}. These methods try to optimize $k$ value in base $k$NN models used in the ensemble. Other ensemble procedures can be found in \cite{aburomman2016novel, gallego2018clustering, zhang2019novel, zhai2017classification, yuan2021using}. Despite of efficient results given by these ensembles on many benchmark problems, they might be affected by the mixed class patterns in the data. Moreover, base learners in these methods are constructed on sub-samples which may ignore discriminative feature(s). Therefore, these procedures need further improvement to overcome such problems.
	
	This paper introduces a novel ensemble method called random projection extended neighbourhood rule (RPExNRule) for $k$NN. This procedure constructs a large number of the extended neighbourhood rule (ExNRule) based classifiers \cite{ali2022k}, each on a randomly projected bootstrap sample and a new data point is predicted by all these models. The final prediction class is obtained using majority voting among the results given by the underlying models. It improves the prediction accuracy in the following steps.
	
	\begin{enumerate}
		\item The $k$ nearest observations to a test sample point in the training data are determined in a step-wise pattern to recognize the true pattern of the data.
		\item The base learners are more random and diverse as they are constructed on randomly projected bootstrap samples.
	\end{enumerate}
	
	\section{The RPExNRule for $k$NN ensemble}
	\label{method}
	Let $\mathcal{L} = (X, Y)_{n\times (p+1)}$ be a training dataset, where $X_{n\times p}$ represnts the feature space with dimension $n \times p$ and $Y$ be a categorical response with two classes i.e. $Y \in (0, 1)$. Draw $B$ bootstrap samples from the training data $\mathcal{L}$ and randomly project the feature space of these samples to dimension $p^\prime \le p$ i.e. $\mathcal{S}_{n\times (p^\prime + 1)}^b$, where $b = 1, 2, 3, \ldots, B$. All these projected samples contain the reduced feature space $\chi_{n\times p^\prime}$ and their corresponding response labels $Y$. 
	
    Suppose $X_{1\times p}^{0}$ is a test sample point and needs to be predicted. In this respect, it will be randomly projected according to the $b^{th}$ random projection i.e. $\chi_{1\times p^\prime}^0$. Then apply the extended neighbourhood rule (ExNRule) \cite{ali2022k} to each random projection and estimate the test data point. The predictions given by all $B$ models of the unseen observation $X_{1\times p}$ are $\hat{Y}^1, \hat{Y}^2, \hat{Y}^3, \ldots , \hat{Y}^{B}$. The final predicted class label of the test sample point is a majority vote of the estimates given by all underlying learners.
	
	\subsection{Mathematical Description}
	\label{sec:Mathematical model}
	Mathematical description of the proposed method is given in the following sub-sections.
	
	\subsubsection{Random Projection}
	The basic idea of random projection method is provided by Johnson-Lindenstrauss lemma \cite{johnson1984extensions}, which is used for dimension reduction \cite{fradkin2003experiments, sulic2010efficient}. This method projects a $p$-dimensional data matrix to a $p^\prime$-dimensional space where $p^\prime \le p$. Mathematically it can be expressed as follows. 
	
	Let $\mathcal{R}_{p \times p^\prime}$ be a random matrix generated from any of the well known distribution such as Gaussian, etc. The elements of $\mathcal{R}_{p \times p^\prime}$ are independently and identically distributed. The given matrix of feature space $X_{n\times p}$ can be projected using matrix multiplication as:
	
	$$\chi_{n\times p^\prime} = X_{n\times p} \mathcal{R}_{p \times p^\prime}$$
	
	The proposed method takes a large number of bootstrap samples and the feature space of each sample is randomly projected. A model based on ExNRule \cite{ali2022k} is fitted to each of these projections.
	
	\subsubsection{The Extended Neighbourhood Rule}
	A distance formula to be used in $b^{th}$ randomly projected bootstrap sample $\mathcal{S}_{n\times (p^\prime + 1)}^b$, where $b = 1, 2, 3, \ldots, B$, to identify the $k$ nearest observations in a step-wise pattern to a test data point $X_{1\times p^\prime}^0$. To make the dimension of the unseen observation according to $b^{th}$ bootstrap sample, it is also randomly projected i.e. $\chi_{1\times p^\prime}^0$. The general formula used for the calculation of distances between data points is given as:
	
	\begin{equation}
		\delta_b(\chi_{1\times p'}^{i-1}, \chi_{1\times p'}^i)_{min}  = \left[\sum_{j=1}^{p'}\|\textit{x}_j^{i-1} - \textit{x}_j^i\|^{q}\right]^{1/q},
		\label{eq1}
	\end{equation}
	where, $i=1, 2, \ldots, k$ and $b=1, 2, \ldots, B$.
	
	In $b^{th}$ randomly projected sample, Equation \ref{eq1} gives the following sequence of minimum distances in a pattern which determines the neighbourhood of the unseen observation.
	
	$$\delta_b(\chi_{1\times p'}^{0}, \chi_{1\times p'}^1)_{min}, \delta_b(\chi_{1\times p'}^{1}, \chi_{1\times p'}^2)_{min},$$
	$$ \delta_b(\chi_{1\times p'}^{2}, \chi_{1\times p'}^3)_{min}, \ldots, \delta_b(\chi_{1\times p'}^{k-1}, \chi_{1\times p'}^k)_{min}.$$
	
	The above sequence determines $\chi_{1\times p'}^{i}$ is the closest data point to $\chi_{1\times p'}^{i-1}$, where, $i = 1 , 2, 3, \ldots, k$. The selected observations are $\chi_{1\times p'}^{1}, \chi_{1\times p'}^{2}, \chi_{1\times p'}^{3}, \ldots, \chi_{1\times p'}^{k}$ and their corresponding classes are $y^1 , y^2, y^3, \ldots,y^k$, respectively. 	Majority voting is used to predict the unseen observation $X_{1\times p}^{0}$ by the $b^{th}$ base learner, i.e., 
	$\hat{Y}^b =$ majority vote of $(y^1, y^2, y^3,\ldots,y^k)$, where, $b=1, 2, 3, \ldots, B $. The final estimation of the class label of that observation is $\hat{Y} =  $ majority vote of $(\hat{Y}^1 , \hat{Y}^2, \hat{Y}^3, \ldots, \hat{Y}^B)$.

	\subsection{Algorithm}
	\label{sec:Algorithm}
	The RPExNRule takes the following steps.
	\begin{enumerate}
		\item Take $B$ bootstrap samples from training data and reduce the original dimension from $p$ to $p^\prime$ using the random projection technique.
		\item Fit extended neighbourhood rule (ExNRule) on each projected sample generated in Step 1.
		\item Predict an unseen observation via all the models fitted in Step 2.
		\item Using majority voting in the results given by all base learners to obtain the final prediction of the unseen observation.
	\end{enumerate}
	
\begin{figure}[h]
		\centering
		\fbox{\includegraphics[width=0.9\textwidth]{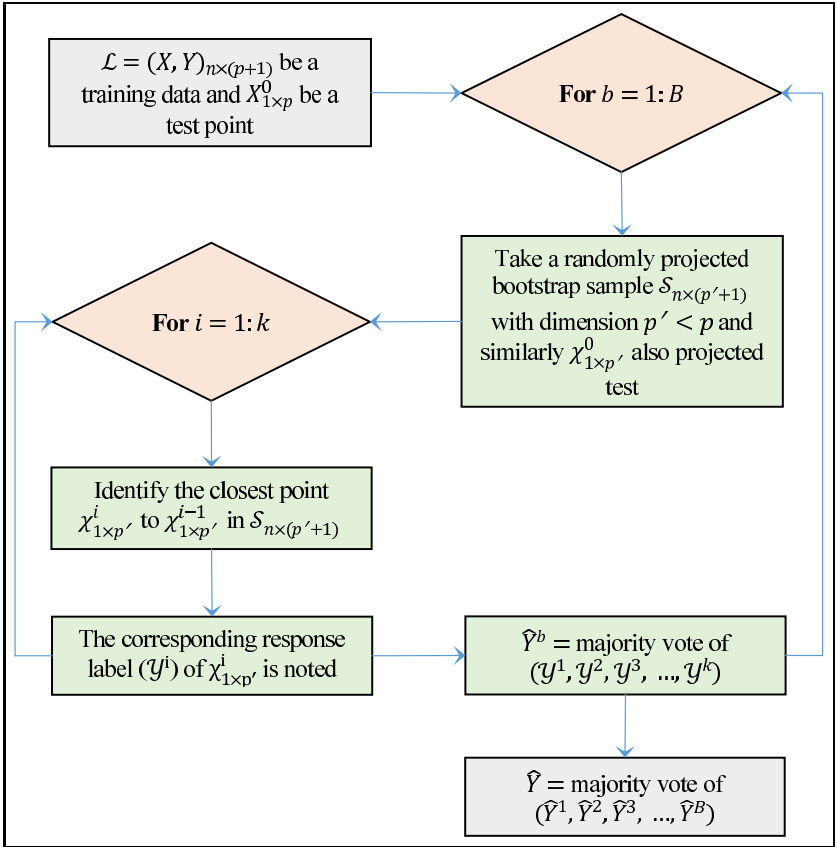}}
		\caption{Flow chart of the random projection extended neighbourhood rule ensemble.}
		\label{Flowchart}
	\end{figure}

	The pseudo code of the proposed RPExNRule method is given in Algorithm \ref{Psudue} and Figure \ref{Flowchart} illustrates the  flowchart of this procedure.
	
	\begin{algorithm} 
		\caption{Pseudo code of the proposed RPExNRule.}
		\begin{algorithmic}[1]
			\State  $\mathcal{L} = (X, Y)_{n\times (p+1)}$ $\leftarrow$ Training data with dimension $n \times (p+1)$
			\State $X_{n \times p} $ $\leftarrow$ Training feature space with $n$ observations and $p$ features;
			\State $Y$ $\leftarrow$ Response vector of $n$ values in the training data;
			\State $X_{1 \times p}^0 $ $\leftarrow$ A test observation to be classified;
			\State $p$ $\leftarrow$ Total number of features in the training data; 
			\State $n$ $\leftarrow$ Total number of observations in training data; 
			\State $B$ $\leftarrow$ Number of randomly projected bootstrap samples each with a reduced dimension $n \times (p^\prime +1)$;
			\State $p'$ $\leftarrow$ Reduced number of features in each projected sample;
			\State $k$ $\leftarrow$ Number of nearest observations to a test point;
			
			\For {$b\gets 1:B$}
			
			\State $ \mathcal{S}_{n \times (p^\prime +1)}^b \gets $ Randomly projected bootstrap sample with reduced feature space $p^\prime$;
			\State $ \chi^0_{1 \times p^\prime} \gets $ Randomly projected test point $X^0_{1 \times p}$ according to training data;
			
			\For {$i\gets 1:k$}
			\State $\chi^i_{1 \times p'} \gets $ Nearest training sample point to $\chi^{i-1}_{1 \times p'}$ in $\mathcal{S}_{n \times (p^\prime +1)}^b$
			\State $y^i \gets$ The corresponding class label
			\EndFor
			
			\State $\hat{Y}^b = $ majority vote of $(y^1 , y^2, y^3, \ldots,y^k)$
			
			\EndFor
			\State $\hat{Y} =  $ majority vote of $(\hat{Y}^1, \hat{Y}^2, \hat{Y}^3, \ldots, \hat{Y}^B)$
		\end{algorithmic}
		\label{Psudue}
	\end{algorithm}
	
	\section{Experiments and results}
	\label{experiment}
	This section provides a detailed discussion on experimental setup, benchmark datasets used for assessment and results given by the proposed method and other classical procedures.
	
	\subsection{Datasets}
	This section gives a detailed discussion on the benchmark and simulation datasets used to evaluate the performance of the proposed method against other classical methods.
	
	\subsubsection{Benchmark datasets}
	To compare the proposed random projection extended neighbourhood rule (RPExNRule) ensemble with other state-of-the-art classifiers, 15 benchmark problems have been used. These datasets are openly available on various repositories. A brief description of these datasets is given in Table \ref{datasets}, showing the data name, number of observations $n$, number of features $p$, class distribution and its source.

	\begin{table}
		\centering
		\caption{Brief description of the datasets along with their corresponding number of features, observations, class-wise distributions and sources.}\label{datasets}%
		\begin{tabular}{@{}lcccl@{}}
			\toprule
			Data    & \multicolumn{1}{l}{p} & \multicolumn{1}{l}{n} & Class Distribution & Sources \\
			\midrule
			KCBI   & 86    & 145   & (60, 85)    & \url{https://www.openml.org/d/1066} \\
			TSVM   & 80    & 156   & (102, 54)   & \url{https://www.openml.org/d/41976} \\
			JEDI   & 8     & 369   & (204, 165)  & \url{https://www.openml.org/d/1048} \\
			WISC   & 32    & 194   & (90, 104)   & \url{https://www.openml.org/d/753} \\
			AR5    & 29    & 36    & (8, 28)     & \url{https://www.openml.org/d/1062} \\	
			ILPD   & 10    & 583   & (167, 415)  & \url{https://www.openml.org/d/1480} \\
			PLRE   & 12    & 182   & (52, 130)   & \url{https://www.openml.org/d/1490} \\
			SLEE   & 7     & 62    & (29, 33)    & \url{https://www.openml.org/d/739} \\
			ECMO   & 9     & 130   & (64, 66)    & \url{https://www.openml.org/d/944} \\
			MC2    & 39    & 161   & (52, 109)   & \url{https://www.openml.org/d/1054} \\
			PREL   & 13    & 315   & (182, 133)  & \url{https://www.openml.org/d/915} \\
			GRDA   & 8     & 155   & (49, 106)   &\url{https://www.openml.org/d/1026} \\
			SLEX   & 9     & 60    & (30, 30)    & \cite{SLEX} \\
			ECOL   & 7     & 336   & (193, 143)  & \cite{ECOL} \\
			SPEC   & 22    & 267   & (55, 212)   & \cite{SPECT} \\
			\bottomrule	
		\end{tabular}
	\end{table}
	
	\subsubsection{Synthetic datasets}
	For assessing the performance of the proposed RPExNRule ensemble, synthetic datasets are generated in three scenarios. Each dataset has 30 features and 200 sample points and a binary response. Half of 200 observations have class $0$ and the remaining half have class $1$. In the feature space 20 out of 30 features are generated from normal distribution with different parameters for each class, while the remaining 10 features are considered as contrived/irrelevant. These features are also generated from the normal distribution with same parameters i.e. $N(\mu = 0, \sigma = 1)$ for both classes. Table \ref{simdata} gives a detailed description of the simulated datasets. In that table the first column gives data IDs, the second and third columns provide the distribution of feature space for class $0$ and class $1$ respectively. The final column gives the distribution of the contrived features in the data.

	\begin{table}[h]
		\centering
		\caption{Brief description of synthetic datasets.}\label{simdata}
		\begin{tabular}{@{}cccc@{}}
			\toprule
			ID & Class $0$ & Class $1$ & Contrived features\\
			\midrule
			$S_1$ & $N(\mu = 1,\sigma = 5)$ & $N(\mu = 0,\sigma = 5)$ & $N(\mu = 0,\sigma = 1)$ \\
			$S_2$ & $N(\mu = 5,\sigma = 5)$ & $N(\mu = 0,\sigma = 5)$ & $N(\mu = 0,\sigma = 1)$ \\
			$S_3$ & $N(\mu = 5,\sigma = 5)$ & $N(\mu = 0,\sigma = 3)$ & $N(\mu = 0,\sigma = 1)$ \\
			\bottomrule
		\end{tabular}
	\end{table}
	
	\subsection{Experimental setup}
	Experimental setup based on 15 benchmark problems provided in Table \ref{datasets} and synthetic datasets given in Table \ref{simdata} is formed as follows. Each of these datasets is randomly divided into two non-overlapping groups i.e. 70\% training and 30\% testing parts of the total number of sample points in the data. The said splitting criteria is repeated 500 times for all datasets. The novel random projection extended neighbourhood rule (RPExNRule) ensemble and other standard classifiers i.e. $k$ nearest neighbours ($k$NN), random $k$ nearest neighbours (R$k$NN), weighted $k$ nearest neighbours (W$k$NN), random forest (RF), optimal trees ensemble (OTE) and support vector machine (SVM) are constructed on the training part and tested on the testing part of each dataset. The final results are provided in Tables \ref{results1}-\ref{results3}.These results are the arithmetic averages of all the results given by 500 splittings.
	
	The proposed RPExNRule ensemble is used with $B = 500$ base learners each on a randomly projected bootstrap sample with reduced number of feature from $p$ to $p^\prime$. The Euclidean distance formula is used to determine the $k = 3$ nearest neighbours in a step-wise pattern. These models are used to predict a test observation using majority voting. The final predicted class of the unseen observation is determined by the second round majority voting of the results given by the base learners. These parameters are used for assessing the performance of the proposed method and all other standard procedures on the benchmark problems. This method is further assessed using $k = 3, 5,$ and $7$ with only $k$NN based classifiers. The proposed method is also compared with $k$NN based techniques on the synthetic datasets using $k = 3$.
	
	The package \texttt{caret} \cite{caret} implemented in R is used to construct $k$NN classifier, while R package \texttt{kknn} \cite{kknn} is used for the fitting of W$k$NN model. The R package \texttt{rknn} \cite{rknn} is used to implement R$k$NN procedure. For the construction of RF and OTE, the R packages \texttt{randomForest} \cite{randomForest} and \texttt{OTE} \cite{OTE} are utilized respectively. To construct SVM model the R package \texttt{kernlab} \cite{kernlab} is used. Moreover, for tuning the hyper-parameter $k$, the R function \texttt{tune.knn} is used which is implemented in the R package \texttt{e1071} \cite{e1071}. For R$k$NN, the hyper-parameters i.e. number of nearest neighbours $k$, number of base learners $r$ etc. are tuned manually in the package \texttt{rknn} \cite{rknn} which is also implemented in R. The R function \texttt{tune.randomForest} is used to tune \texttt{ntree}, \texttt{nodesize} and \texttt{mtry} in R package \texttt{e1071} \cite{e1071}. The same tuned parameters are used for OTE procedure in R package \texttt{OTE} \cite{OTE}. The R package \texttt{kernlab} \cite{kernlab} is used for SVM classifier with linear kernel and other  default parameter.
	
	\subsection{Results}
	Table \ref{results1} gives the results for all considered datasets which shows that the proposed RPExNRule has outperformed the other standard methods in majority of the cases. The novel RPExNRule is providing maximum accuracy compared to the other state-of-the-art methods on 12 datasets. The standard R$k$NN outperformed on 1 dataset, while RF gives high accuracy on 2 datasets. The classical $k$NN, W$k$NN, OTE and SVM procedures do not give satisfactory results on any of the datasets. 
	
\begin{figure}[H]
		\centering
		\includegraphics[width=0.9\textwidth]{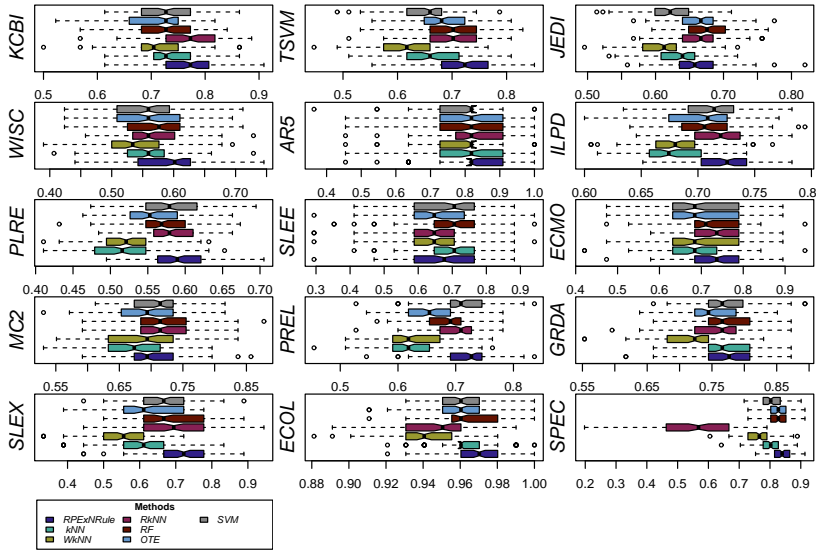}
		\caption{This figure shows boxplots of accuracy of the proposed RPExNRule and other classifiers for all datasets.}
		\label{Acc15}
	\end{figure}

\begin{figure}[H]
		\centering
		\includegraphics[width=0.9\textwidth]{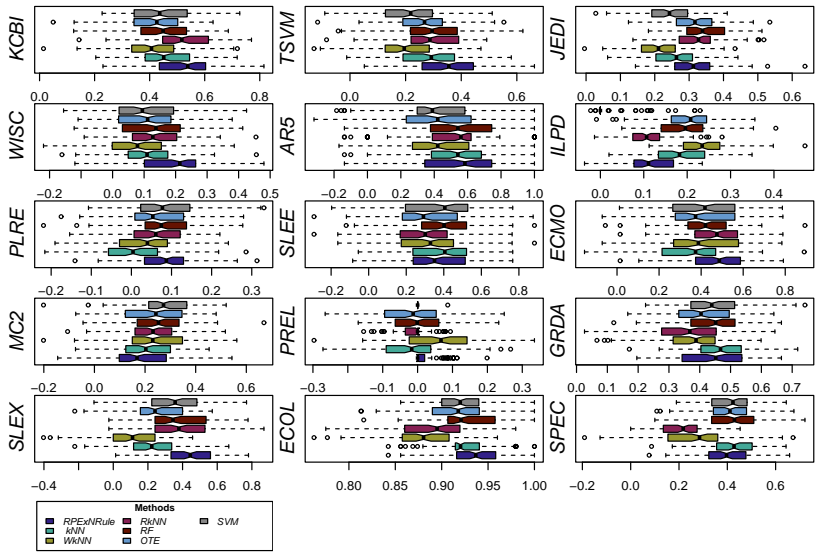}
		\caption{This figure shows boxplots of Cohen's kappa of the proposed RPExNRule and other classifiers for all datasets.}
		\label{Kappa15}
	\end{figure}

\begin{figure}[H]
		\centering
		\includegraphics[width=0.9\textwidth]{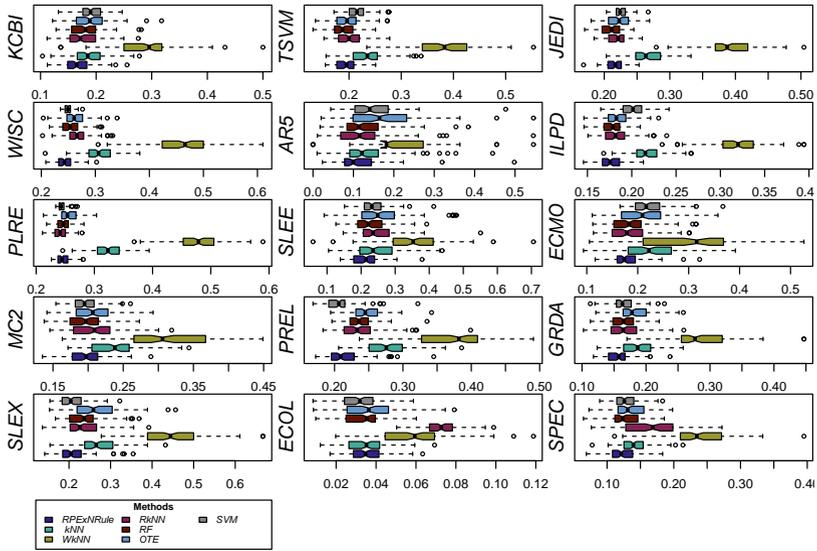}
		\caption{This figure shows boxplots of Brier score (BS) of the proposed RPExNRule and other classifiers for all datasets.}
		\label{BS15}
	\end{figure}

	In case of Cohen's kappa, the novel RPExNRule provides optimal results on 8 datasets, while $k$NN method gives maximum kappa values on 2 out of the total datasets. The W$k$NN procedure outperformed on 2 datasets, while R$k$NN and OTE did not give optimal result on any of the datasets. The RF and SVM also give optimal results each on 2 datasets. Similarly, the proposed RPExNRule ensemble outperformed all the other standard methods in terms of Brier score (BS). Furthermore, boxplots of the results are also constructed and provided in the Figures \ref{Acc15}, \ref{Kappa15} and \ref{BS15}.
	
\begin{figure}[H]
		\centering
		\includegraphics[width=0.9\textwidth]{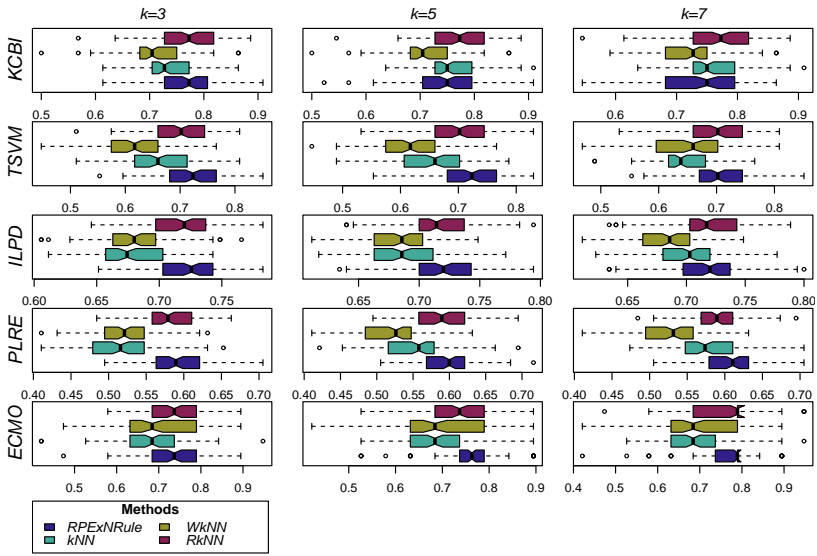}
		\caption{This figure shows boxplots of accuracy of the proposed RPExNRule and $k$NN based classifiers for different $k$ values.}
		\label{Acc5}
	\end{figure}

\begin{figure}[H]
		\centering
		\includegraphics[width=0.9\textwidth]{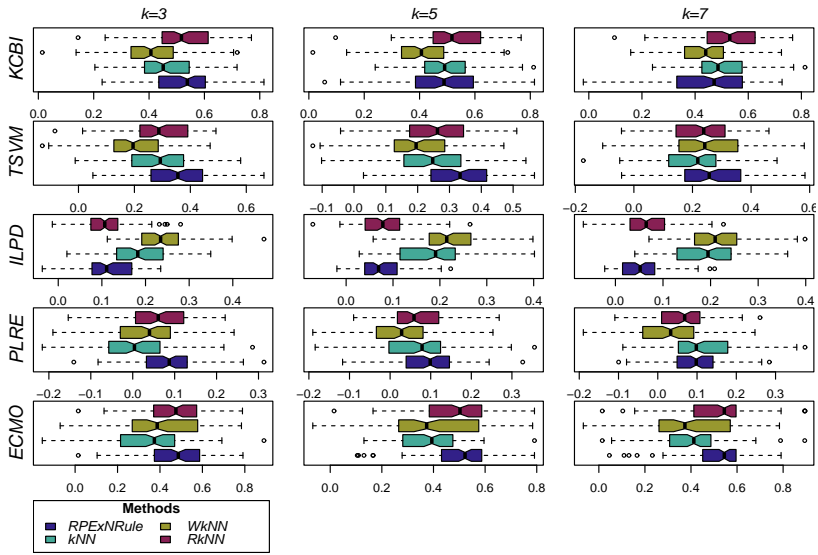}
		\caption{This figure shows boxplots of Cohen's kappa of the proposed RPExNRule and $k$NN based classifiers for different $k$ values.}
		\label{Kappa5}
	\end{figure}

\begin{figure}[H]
		\centering
		\includegraphics[width=0.9\textwidth]{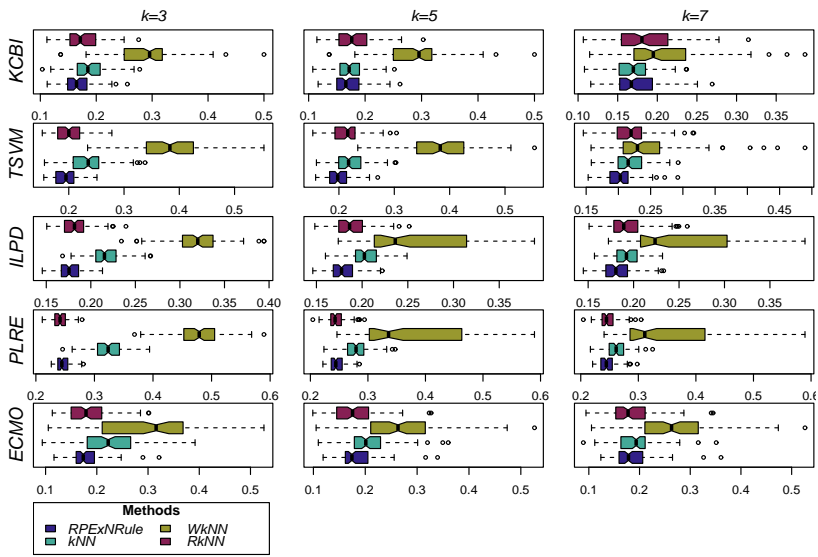}
		\caption{This figure shows boxplots of Brier score (BS) of the proposed RPExNRule and $k$NN based classifiers for different $k$ values.}
		\label{BS5}
	\end{figure}

	Moreover, performance of the proposed RPExNRule and other standard $k$NN based procedures (i.e. $k$NN, W$k$NN and R$k$NN) is also assessed by using various values of $k$ (i.e. $k = 3, 5, 7$). The results computed for 5 datasets are given in Table \ref{results2}. It shows that the novel method has outperformed the other models in majority of the cases. The results also reveal that the proposed method is more stable for different $k$ values as compared to other $k$NN base procedures. Furthermore, boxplots of these results are given in Figures \ref{Acc5}, \ref{Kappa5} and \ref{BS5}.
	
\begin{figure}[H]
		\centering
		\includegraphics[width=0.9\textwidth]{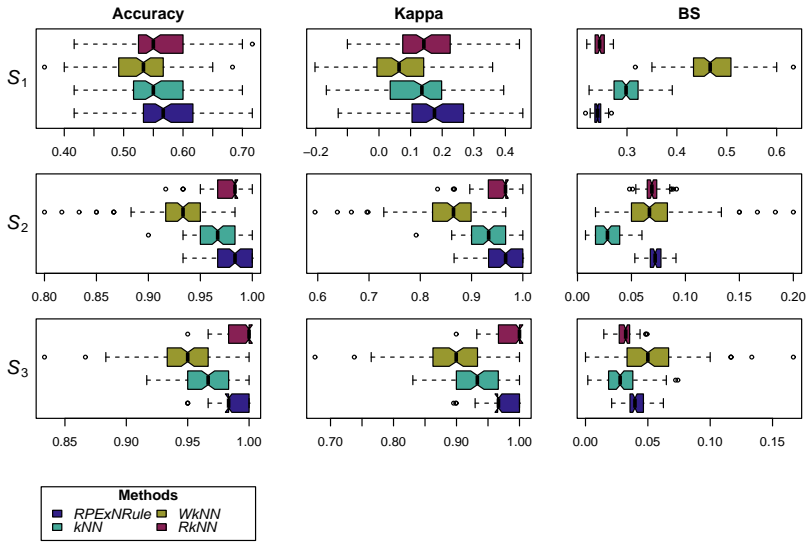}
		\caption{This figure shows boxplots of accuracy, Cohen's kappa and Brier score (BS) of the proposed RPExNRule and $k$NN based classifiers for synthetic datasets.}
		\label{sim}
	\end{figure}

	The performance of the proposed method is also assessed on synthetic datasets in comparison with the other classical $k$NN based classifiers. The results are given in Table \ref{results3}, which shows optimal performance in majority of the given scenarios. Table \ref{results3} reveals that the proposed RPExNRule ensemble has optimal results when observations of both classes are mixed up and have high variation. Moreover, Figure \ref{sim} gives the boxplots of the results computed via RPExNRule and other classical methods on the synthetic datasets.

	\begin{sidewaystable}
		\sidewaystablefn%
		\begin{center}
			\begin{minipage}{\textheight}
				\setlength{\tabcolsep}{2pt}
				\fontsize{9.5}{9.5}\selectfont
				\renewcommand{\arraystretch}{1}
				\caption{This table shows accuracy, Cohen's kappa and Brier score (BS) of the proposed RPExNRule and other classifiers on all benchmark datasets.}\label{results1}
				\begin{tabular}{llcccccccccccccccc}
					\toprule
					\multirow{2}{*}{Metrics} & \multicolumn{1}{c}{\multirow{2}{*}{Methods}} & \multicolumn{15}{c}{Datasets} & \multicolumn{1}{c}{\multirow{2}{*}{Mean}} \\
					\cmidrule(lr){3-17}
					& \multicolumn{1}{c}{} & KCBI & TSVM & JEDI & WISC & AR5 & ILPD & PLRE & SLEE & ECMO & MC2 & PREL & GRDA & SLEX & ECOL & SPEC &  \multicolumn{1}{c}{} \\
					\midrule
					
					\multirow{7}{*}{Accuracy} & OExNRule & \textbf{0.766} & \textbf{0.724} & 0.663 & \textbf{0.594} & \textbf{0.836} & \textbf{0.722} & \textbf{0.591} & 0.675 & \textbf{0.737} & 0.701 & \textbf{0.717} & \textbf{0.774} & \textbf{0.723} & \textbf{0.967} & \textbf{0.838} & \textbf{0.735} \\
					& $k$NN & 0.741 & 0.666 & 0.632 & 0.556 & 0.817 & 0.678 & 0.515 & 0.680 & 0.677 & 0.675 & 0.625 & 0.770 & 0.608 & 0.965 & 0.803 & 0.694 \\
					& W$k$NN & 0.709 & 0.620 & 0.608 & 0.538 & 0.785 & 0.682 & 0.521 & 0.661 & 0.707 & 0.686 & 0.623 & 0.720 & 0.557 & 0.943 & 0.761 & 0.675 \\
					& R$k$NN & 0.762 & 0.701 & 0.665 & 0.568 & 0.829 & 0.718 & 0.582 & 0.640 & 0.730 & \textbf{0.715} & 0.701 & 0.754 & 0.691 & 0.947 & 0.541 & 0.703 \\
					& RF & 0.723 & 0.698 & \textbf{0.677} & 0.567 & 0.824 & 0.707 & 0.570 & \textbf{0.684} & 0.710 & 0.711 & 0.679 & 0.768 & 0.686 & 0.965 & 0.825 & 0.720 \\
					& OTE & 0.711 & 0.682 & 0.663 & 0.561 & 0.790 & 0.702 & 0.559 & 0.650 & 0.702 & 0.694 & 0.653 & 0.747 & 0.628 & 0.959 & 0.824 & 0.702 \\
					& SVM & 0.734 & 0.641 & 0.623 & 0.551 & 0.785 & 0.710 & 0.580 & 0.679 & 0.698 & 0.706 & 0.713 & 0.772 & 0.679 & 0.963 & 0.807 & 0.709 \\
					&  &  &  &  &  &  &  &  &  &  &  &  &  &  &  &  &  \\
					\multirow{7}{*}{Kappa} & OExNRule & \textbf{0.520} & \textbf{0.352} & 0.314 & \textbf{0.189} & \textbf{0.527} & 0.116 & \textbf{0.083} & 0.358 & \textbf{0.477} & 0.182 & 0.023 & 0.440 & \textbf{0.452} & \textbf{0.931} & 0.395 & \textbf{0.357} \\
					& $k$NN & 0.465 & 0.283 & 0.254 & 0.113 & 0.511 & 0.186 & 0.001 & 0.363 & 0.355 & 0.199 & -0.027 & \textbf{0.467} & 0.221 & 0.927 & \textbf{0.426} & 0.316 \\
					& W$k$NN & 0.406 & 0.201 & 0.211 & 0.076 & 0.429 & \textbf{0.238} & 0.033 & 0.322 & 0.412 & 0.241 & \textbf{0.063} & 0.374 & 0.125 & 0.882 & 0.262 & 0.285 \\
					& R$k$NN & 0.516 & 0.289 & 0.320 & 0.132 & 0.519 & 0.108 & 0.064 & 0.288 & 0.461 & 0.232 & -0.012 & 0.363 & 0.384 & 0.891 & 0.206 & 0.317 \\
					& RF & 0.441 & 0.299 & \textbf{0.347} & 0.132 & 0.504 & 0.197 & 0.081 & \textbf{0.378} & 0.422 & 0.254 & 0.003 & 0.447 & 0.373 & 0.927 & 0.421 & 0.348 \\
					& OTE & 0.418 & 0.265 & 0.316 & 0.117 & 0.426 & 0.201 & 0.064 & 0.309 & 0.402 & 0.235 & -0.013 & 0.410 & 0.258 & 0.916 & 0.408 & 0.315 \\
					& SVM & 0.447 & 0.212 & 0.241 & 0.101 & 0.419 & 0.018 & \textbf{0.083} & 0.363 & 0.397 & \textbf{0.280} & 0.001 & 0.447 & 0.363 & 0.924 & 0.407 & 0.314 \\
					&  &  &  &  &  &  &  &  &  &  &  &  &  &  &  &  &  \\
					\multirow{7}{*}{BS} & OExNRule & \textbf{0.167} & 0.196 & 0.217 & \textbf{0.244} & \textbf{0.117} & \textbf{0.177} & 0.247 & \textbf{0.214} & \textbf{0.180} & 0.197 & 0.215 & \textbf{0.157} & 0.210 & 0.035 & \textbf{0.124} & \textbf{0.180} \\
					& $k$NN & 0.186 & 0.236 & 0.266 & 0.308 & 0.133 & 0.218 & 0.324 & 0.243 & 0.227 & 0.237 & 0.278 & 0.186 & 0.271 & 0.034 & 0.140 & 0.219 \\
					& W$k$NN & 0.291 & 0.380 & 0.392 & 0.462 & 0.215 & 0.318 & 0.479 & 0.339 & 0.293 & 0.314 & 0.377 & 0.280 & 0.443 & 0.057 & 0.239 & 0.325 \\
					& R$k$NN & 0.176 & 0.202 & 0.219 & 0.267 & 0.122 & 0.183 & \textbf{0.242} & 0.244 & 0.183 & 0.209 & 0.237 & 0.168 & 0.234 & 0.073 & 0.166 & 0.195 \\
					& RF & 0.180 & \textbf{0.189} & \textbf{0.211} & 0.255 & 0.126 & 0.178 & 0.247 & 0.227 & 0.186 & \textbf{0.196} & 0.238 & 0.166 & 0.236 & 0.034 & 0.128 & 0.186 \\
					& OTE & 0.191 & 0.197 & 0.222 & 0.263 & 0.178 & 0.184 & 0.256 & 0.260 & 0.208 & 0.206 & 0.247 & 0.182 & 0.263 & 0.036 & 0.137 & 0.202 \\
					& SVM & 0.193 & 0.216 & 0.226 & 0.250 & 0.148 & 0.200 & 0.244 & 0.238 & 0.222 & 0.198 & \textbf{0.208} & 0.165 & \textbf{0.209} & \textbf{0.032} & 0.128 & 0.192\\
					
					\bottomrule
				\end{tabular}%
			\end{minipage}
		\end{center}
	\end{sidewaystable}

	\begin{sidewaystable}
		\sidewaystablefn%
		\begin{center}
			\begin{minipage}{\textheight}
				\setlength{\tabcolsep}{2pt}
				\fontsize{9.5}{9.5}\selectfont
				\renewcommand{\arraystretch}{1}
				\caption{This table shows accuracy, Cohen's kappa and Brier score (BS) of the proposed RPExNRule and $k$NN based classifiers for different values of $k$ on five benchmark datasets.}\label{results2}
				
				\begin{tabular}{llcccccccccccccccc}
					\toprule
					\multirow{3}{*}{Metrics} & \multicolumn{1}{c}{\multirow{3}{*}{Methods}} & \multicolumn{15}{c}{Datasets} & \multicolumn{1}{c}{\multirow{3}{*}{Mean}} \\
					\cline{3-17}
					& \multicolumn{1}{c}{} & \multicolumn{3}{c}{$KCBI$} & \multicolumn{3}{c}{$TSVM$} & \multicolumn{3}{c}{$JEDI$} & \multicolumn{3}{c}{$WISC$} & \multicolumn{3}{c}{$AR$} & \multicolumn{1}{c}{} \\ \cmidrule(lr){3-5}\cmidrule(lr){9-11} \cmidrule(lr){15-17} \cmidrule(lr){6-8} \cmidrule(lr){12-14}
					& \multicolumn{1}{c}{} & $k = 3$ & $k = 5$ & $k = 7$ & $k = 3$ & $k = 5$ & $k = 7$ & $k = 3$ & $k = 5$ & $k = 7$ & $k = 3$ & $k = 5$ & $k = 7$ & $k = 3$ & $k = 5$ & $k = 7$ & \multicolumn{1}{c}{} \\
					\midrule
					\multirow{4}{*}{Accuracy} & RPEXNRule & \textbf{0.766} & 0.751 & 0.736 & \textbf{0.724} & \textbf{0.724} & \textbf{0.712} & \textbf{0.722} & \textbf{0.720} & \textbf{0.719} & \textbf{0.591} & \textbf{0.600} & \textbf{0.605} & \textbf{0.737} & \textbf{0.755} & \textbf{0.761} & \textbf{0.708} \\
					& $k$NN & 0.741 & 0.757 & 0.758 & 0.666 & 0.657 & 0.650 & 0.678 & 0.688 & 0.699 & 0.515 & 0.550 & 0.576 & 0.677 & 0.693 & 0.702 & 0.667 \\
					& W$k$NN & 0.709 & 0.709 & 0.727 & 0.620 & 0.620 & 0.649 & 0.682 & 0.683 & 0.683 & 0.521 & 0.519 & 0.525 & 0.707 & 0.702 & 0.700 & 0.650 \\
					& R$k$NN & 0.762 & \textbf{0.765} & \textbf{0.765} & 0.701 & 0.696 & 0.695 & 0.718 & 0.717 & 0.718 & 0.582 & 0.586 & 0.589 & 0.730 & 0.747 & 0.757 & 0.702 \\
					\multicolumn{1}{l}{} &  &  &  &  &  &  &  &  &  &  &  &  &  &  &  &  &  \\
					\multirow{4}{*}{Kappa} & RPEXNRule & \textbf{0.520} & 0.486 & 0.450 & \textbf{0.352} & \textbf{0.327} & \textbf{0.267} & 0.116 & 0.077 & 0.057 & \textbf{0.083} & \textbf{0.094} & 0.098 & \textbf{0.477} & \textbf{0.513} & \textbf{0.526} & \textbf{0.296} \\
					& $k$NN & 0.465 & 0.498 & 0.501 & 0.283 & 0.245 & 0.205 & 0.186 & 0.184 & 0.195 & 0.001 & 0.063 & \textbf{0.109} & 0.355 & 0.387 & 0.409 & 0.272 \\
					& W$k$NN & 0.406 & 0.406 & 0.441 & 0.201 & 0.201 & 0.242 & \textbf{0.238} & \textbf{0.221} & \textbf{0.215} & 0.033 & 0.021{\tiny } & 0.029 & 0.412 & 0.404 & 0.401 & 0.258 \\
					& R$k$NN & 0.516 & \textbf{0.524} & \textbf{0.525} & 0.289 & 0.258 & 0.234 & 0.108 & 0.079 & 0.071 & 0.064 & 0.064 & 0.064 & 0.461 & 0.500 & 0.520 & 0.285 \\
					\multicolumn{1}{l}{} &  &  &  &  &  &  &  &  &  &  &  &  &  &  &  &  &  \\
					\multirow{4}{*}{BS} & RPEXNRule & \textbf{0.167} & \textbf{0.170} & 0.173 & \textbf{0.196} & \textbf{0.199} & \textbf{0.203} & \textbf{0.177} & \textbf{0.180} & \textbf{0.183} & 0.247 & 0.247 & \textbf{0.247} & \textbf{0.180} & 0.183 & 0.186 & \textbf{0.196} \\
					& $k$NN & 0.186 & 0.171 & \textbf{0.169} & 0.236 & 0.221 & 0.219 & 0.218 & 0.203 & 0.193 & 0.324 & 0.280 & 0.262 & 0.227 & 0.207 & 0.193 & 0.221 \\
					& W$k$NN & 0.291 & 0.291 & 0.207 & 0.380 & 0.380 & 0.244 & 0.318 & 0.258 & 0.250 & 0.479 & 0.377 & 0.349 & 0.293 & 0.277 & 0.270 & 0.311 \\
					& R$k$NN & 0.176 & 0.181 & 0.185 & 0.202 & 0.214 & 0.221 & 0.183 & 0.189 & 0.194 & \textbf{0.242} & \textbf{0.245} & 0.248 & 0.183 & \textbf{0.179} & \textbf{0.185} & 0.202 \\
					\bottomrule
				\end{tabular}%
			\end{minipage}
		\end{center}
	\end{sidewaystable}
	
	\begin{table}[h]
		\begin{center}
			\caption{This table shows accuracy, Cohen's kappa and Brier score (BS) of the proposed RPExNRule and $k$NN based classifiers on synthetic datasets.}	\label{results3}
			
			\begin{tabular}{llccc}
				\toprule
				\multirow{2}{*}{Metrics} & \multicolumn{1}{c}{\multirow{2}{*}{Methods}} & \multicolumn{3}{c}{Scenario} \\ 
				\cmidrule(lr){3-5}
				& \multicolumn{1}{c}{} & $S_1$ & $S_2$ & $S_3$ \\ 
				\midrule
				\multirow{4}{*}{Accuracy} & RPExNRule & \textbf{0.574} & \textbf{0.982} & 0.987 \\
				& $k$NN & 0.555 & 0.971 & 0.966 \\
				& W$k$NN & 0.530 & 0.927 & 0.948 \\
				& R$k$NN & 0.560 & 0.978 & \textbf{0.994} \\
				\multicolumn{1}{l}{} &  &  &  &  \\
				\multirow{4}{*}{Kappa} & RPExNRule & \textbf{0.184} & \textbf{0.964} & 0.973 \\
				& $k$NN & 0.118 & 0.941 & 0.931 \\
				& W$k$NN & 0.064 & 0.852 & 0.895 \\
				& R$k$NN & 0.152 & 0.955 & \textbf{0.987} \\
				\multicolumn{1}{l}{} &  &  &  &  \\
				\multirow{4}{*}{BS} & RPExNRule & \textbf{0.242} & 0.072 & 0.041 \\
				& $k$NN & 0.300 & \textbf{0.029} & 0.029 \\
				& W$k$NN & 0.470 & 0.073 & 0.052 \\
				& R$k$NN & 0.245 & 0.069 & \textbf{0.032}\\
				\bottomrule
			\end{tabular}
		\end{center}
	\end{table}

	
	
	\section{Conclusion}
	\label{conclusion}
	This paper proposed an ensemble method using extended neighbourhood rule (ExNRule) for constructing base $k$NN models, each on a randomly projected bootstrap sample. The $k$ nearest observations to an unseen data point are identified in a $k$ steps pattern. In the first step, it determines the nearest observation to the test sample point. In the second step, it searches for another observation which is closest to the previously selected sample point. This process is continued until the desired $k$ observations are obtained in the neighbourhood. Each base learner uses majority voting in the response classes of the $k$ nearest observations to predict the class of the unseen observation. The final estimated class of the test observation is a second round majority vote of the results obtained from the base learners. The performance of the proposed method is validated on 15 benchmark problems and 3 synthetic datasets. The results are compared with classical $k$NN, W$k$NN, R$k$NN, RF, OTE and SVM classifiers. The classification accuracy, Cohen's kappa and Brier score (BS) are used as performance metrics.
	
	The basic intuition behind the performance of the proposed ensemble is the step-wise selection of the $k$ closest observations and randomly projecting the feature space.	The standard $k$NN based ensembles ignore influential features in many cases which leads to its poor prediction performance. Moreover, these procedures fail to perform well in a case where unseen data observations follow a pattern of the nearest sample points in the training data that lie on a certain path not enclosed by the given sphere of neighbourhood. To overcome these issues, the proposed RPExNRule ensemble uses the extended neighbourhood rule (ExNRule) to fit $k$NN as base classifiers on randomly projected bootstrap samples. The results also reveal that the proposed method is less sensitive to the parameter $k$ as compare to the other $k$NN base classifiers. Moreover, base models are constructed on randomly projected bootstrap samples, drawn from training data, which reduces the dimension without any information loss and ensure diversity due to randomization in the base learners. 
	
	The proposed ensemble combines a large number of $B$ ExNRule base learners each constructed on a randomly projected bootstrap sample and the closest observation is selected in a step-wise manner, which makes the procedure time-consuming and laborious. To solve this problem, it is required to use R package \texttt{parallel} \cite{parallel} to parallelize steps 8-16 of Algorithm \ref{Psudue}. The performance can further be improved by selecting a suitable distance formula. Feature selection method such as \cite{tawhid2020feature, hamraz2021robust, hashemi2022ensemble, wang2022sparse}, could be used for further improvement in the performance of the proposed method.

\bibliography{mybibfile.bib}
\end{document}